\pdfoutput=1
\let\accentvec\vec
\documentclass{llncs}
     
     \let\vec\accentvec
\usepackage{amsmath}
\usepackage{amssymb}
\usepackage{mathtools}
\usepackage{url,graphicx}

\begin{document}
%
%
\pagestyle{headings}  
%

\mainmatter 
\title{Document Clustering Games in Static and Dynamic Scenarios}

\titlerunning{Document Clustering Games}  
%
\author{Rocco Tripodi\inst{1} \and Marcello Pelillo\inst{1,2}}
\authorrunning{R. Tripodi and M. Pelillo} 
%
\tocauthor{Rocco Tripodi and Marcello Pelillo}
\institute{ECLT, Ca' Foscari University, Ca' Minich, Venice, Italy,\\
\and
DAIS, Ca' Foscari University, Via Torino, Venice, Italy.\\
 \{rocco.tripodi,pelillo\}@unive.it}

\maketitle              

\begin{abstract}%
In this work we propose a game theoretic model for document clustering. Each
document to be clustered is represented as a player and each cluster as a
strategy. The players receive a reward interacting with other players that they
try to maximize choosing their best strategies. The geometry of the data is
modeled with a weighted graph that encodes the pairwise similarity among
documents, so that similar players are constrained to choose similar strategies,
updating their strategy preferences at each iteration of the games. We used
different approaches to find the prototypical elements of the clusters and with
this information we divided the players into two disjoint sets, one collecting
players with a definite strategy and the other one collecting players that try
to learn from others the correct strategy to play. The latter set of players can
be considered as new data points that have to be clustered according to previous
information. This representation is useful in scenarios in which the data are
streamed continuously. The evaluation of the system was conducted on 13 document
datasets using different settings. It shows that the proposed method performs
well compared to different document clustering algorithms\footnote{This paper will be published in the series Lecture Notes in Computer Science (LNCS) published by Springer, containing the ICPRAM 2016 best papers.}.
\end{abstract}

\section{Introduction} \label{sec:introduction}%
Document clustering is a particular kind of clustering that involves textual
data. It can be employed to organize tweets
\cite{sankaranarayanan2009twitterstand}, news \cite{bharat2009methods}, novels
\cite{ardanuy2014structure} and medical documents \cite{dhillon2001co}. It is a
fundamental task in text mining and have different applications in document
organization and language modeling \cite{manning2008introduction}.

State-of-the-art algorithms designed for this task are based on generative
models \cite{zhong2005generative}, graph models
\cite{zhao2005hierarchical,tagarelli2013document} and matrix factorization
techniques \cite{xu2003document,pompili2014two}. Generative models and topic
models \cite{Blei:2003:LDA:944919.944937} aim at finding the underlying
distribution that created the set of data objects, observing the sequences of
objects and features. One problem with these approaches is the
conditional-independence assumption that does not hold for textual data and in
particular for streaming documents. In fact, streamed documents such as mails,
tweets or news can be generated in response to past events, creating topics and
stories that evolve over time.

CLUTO is a popular graph-based algorithm for document clustering
\cite{zhao2004empirical}. It employs a graph to organize the documents and
different criterion functions to partition this graph into a predefined number
of clusters. The problem with partitional approaches is that these approaches
require to know in advance the number of clusters into which the data points
have to be divided. A problem that can be restrictive in real applications and 
in particular on streaming data.

Matrix factorization algorithms, such as Non-negative Matrix Factorization (NMF)
\cite{lee1999learning,ding2006nonnegative}, assume that words that occur
together can represent the features that characterize a clusters. Ding et al.
\cite{ding2006nonnegative} demonstrated the equivalence between NMF and
Probabilistic Latent Semantic Indexing, a popular technique for document
clustering. Also with these approaches it is required to know in advance the
number of clusters into which the data have to be organized.

A general problem, common to all these approaches, concerns the temporal
dimension. In fact, for these approaches it is difficult to deal with streaming
datasets. A non trivial problem, since in many real world applications documents
are streamed continuously. This problem is due to the fact that these approaches
operate on a dataset as a whole and need to be recomputed if the dataset
changes. It can be relevant also in case of huge static datasets, because of
scalability issues \cite{DBLP:books/crc/aggarwal13/Aggarwal13a}. In these 
contexts an incremental algorithm would be preferable, since with this approach 
it is possible to cluster the data sequencially.

 With our approach we try to overcome this problem. We cluster part of the data
 producing small clusters that at the beginning of the process can be considered
 as cluster representative. Then we cluster new instances according to this
 information. With our approach is also possible deal with situations in which
 the number of clusters is unknown, a common situation in real world
 applications. The clustering of new instances is defined as a game, in which
 there are labeled players (from an initial clustering), which always play the
 strategy associated to their cluster and unlabeled players that learn their
 strategy playing the games iteratively and obtaining a feedback from the
 strategy that their co-players are adopting.
 

In contrast to other stream clustering algorithm our approach is not based only
on proximity relations, such as in methods based on partitioning representatives
\cite{aggarwal2014data}. With these approaches the cluster membership of new
data points is defined selecting the cluster of their closest representative.
With our approach the cluster membership emerges dynamically from the
interactions of the players and all the neighbors of a new data point contribute
in different proportion to the final cluster assignment. It does not consider
only local information to cluster new data points but find solutions that are
globally consistent. In fact, if we consider only local information the cluster
membership of a point in between two or more clusters could be arbitrary.
 
The rest of this contribution is organized as follows. In the next Section, we
briefly introduce the basic concepts of classical game theory and evolutionary
game theory that we used in our framework; for a more detailed analysis of these
topics the reader is referred to
\cite{weibull1997evolutionary,leyton2008essentials,sandholm2010population}. Then
we introduce the \emph{dominant set} clustering algorithm
\cite{pavan2007dominant,rota2013game} that we used in part
of our experiments to find the initial clustering of the data. In Section
\ref{sec:docClustGames} we describe our model and in the last section we present
the evaluation of our approach in different scenarios. First we use it to
cluster static datasets and then, in Section \ref{sec:expStream}, we present the
evaluation of our method on streaming data. This part extends our previous work
\cite{icpram16} and demonstrates that the proposed framework can be used in
different scenarios with good performances.
\section{Game Theory} \label{sec:GT}%
Game theory was introduced by Von Neumann and Morgenstern \cite{von1944theory}.
Their idea was to develop a mathematical framework able to model the essentials
of decision making in interactive situations. In its \textit{normal-form}
representation, which is the one we use in this work, it consists of a finite
set of players $I=\{1,..,n\}$, a set of pure strategies, $S_i=\{s_1, ...,
s_m\}$, and a utility function $u_i : S_1 \times ... \times S_n \rightarrow
\mathbb{R}$ that associates strategies to payoffs; $n$ is the number of players
and $m$ the number of pure strategies. The games are played among two different
players and each of them have to select a strategy. The outcome of a game
depends on the combination of strategies (strategy profile) played at the same
time by the players involved in it, not just on the single strategy chosen by a
player. For example we can consider the following payoff matrix,

\begin{table} \begin{center} \begin{tabular}{ l | c c } $P_1 \backslash P_2$ &
strategy 1 & strategy 2 \\ \hline strategy 1 & -5,-5 & 0,-6 \\ strategy 2 & -6,0
& -1,-1 \end{tabular} \end{center} \caption{\label{tab:prisoner} The payoff
matrix of the prisoner's dilemma game. } \end{table}

\noindent where, for example, player $1$ get $-5$ when he chooses strategy $1$
and player $2$ chooses strategy $1$. Furthermore, in \emph{non-cooperative
games} the players choose their strategies independently, considering what the
other players can play and try to find the best strategy profile to employ in a
game. 

An important assumption in game theory is that the players try to maximize their
utility in the games ($u_i$), selecting the strategies that can give the highest
payoff, considering what strategies the other player can employ. The players try
to find the strategies that are better than others regardless what the other
player does. These strategies are called \emph{strictly dominant} and can occur
if and only if:

\begin{equation}\label{eq:domstrat}
u(s_i^*,s_{-i})>u_i(s_i,s_{-i}), \forall s_{-i} \in S_{-i}
\end{equation}

\noindent where $s_{-i}$ denotes the strategy chosen by the other player(s).

The key concept of game theory is the Nash equilibrium that is used to predict
the outcome of a strategic interaction. It can be defined as those strategy
profiles in which no player has the incentive to unilaterally deviate from it,
because there is no way to do increment the payoff. The strategies in a Nash
equilibrium are best responses to all other strategies in the game, which means
that they give the most favorable outcome for a player, given other players'
strategies.

The players can play \emph{mixed strategies}, which are probability
distributions over pure strategies. In this context, the players select a
strategy with a certain probability. A mixed strategy set can be defined as a
vector $x=(x^{1},\ldots,x^{m} )$, where $m$ is the number of pure strategies and
each component $x^{h}$ denotes the probability that a particular player select
its $h$th pure strategy. Each player has a strategy set that is defined as a
standard simplex:

\begin{equation}\label{eq:simplex}
 \Delta = \Big\{ x \in \mathbb{R} : \sum_{h=1}^m x^{h} = 1,\text{ and } x^{h} \geq 0 \textit{ for all } h \Big\}.
\end{equation}

\noindent A mixed strategy set corresponds to a point on the simplex $\delta$,
whose corners represent pure strategies.

A strategy profile can be defined as a pair $(p,q)$ where $p \in \Delta_i$ and
$q \in \Delta_j$. The payoff of this strategy profile is computed as:

\begin{equation}\label{eq:gtpayoff}
  u_i(p,q)=p \cdot A_i q \textit{ , } u_j(p,q)=q \cdot A_j p,
\end{equation}

\noindent where $A_i$ and $A_j$ are the payoff matrices of player $i$ and $j$
respectively. The Nash equilibrium within this setting can be computed in the
same way it is computed in pure strategies. In this case, it consists in a pair
of mixed strategies such that each one is a best response to the other.

To overcome some limitations of traditional game theory, such as the
hyper-rationality imposed on the players, a dynamic version of game theory was
introduced. It was proposed by John Maynard Smith and George Price
\cite{smith1973conflict}, as evolutionary game theory. Within this
framework the games are not static and are played repeatedly. This reflect real
life situations, in which the choices change according to past experience.
Furthermore, players can change a behavior according to heuristics or social
norms \cite{szabo2007evolutionary}. In this context, players make a choice that
maximizes their payoffs, balancing cost against benefits
\cite{okasha2012evolution}.

From a machine learning perspective this process can be seen as an
\emph{inductive learning} process, in which agents play the games repeatedly and
at each iteration of the system they update their beliefs on the strategy to
take. The update is done considering what strategy has been effective and what
has not in previous games. With this informatioin, derived from the observation
of the payoffs obtained by each strategy, the players can select the strategy
with higher payoff.

The strategy space of each players is defined as a mixed strategy profile $x_i$
and the mixed strategy space of the game is given by the Cartesian product of
all the players' strategy space:

\begin{equation}
  \Theta = \times_{i \in I} \Delta_i.
\end{equation}

\noindent The expected payoff of a strategy $e^h$ in a single game is calculated
as in mixed strategies (see Equation \ref{eq:gtpayoff}) but, in evolutionary
game theory, the final payoff of each player is the sum of all the partial
payoffs obtained during an iteration. The payoff corresponding to a single
strategy is computed as:

\begin{equation}\label{eq:singlePayoff}
 u_i(e_i^h) = \sum_{j=1}^n(A_{ij} x_j)^h
\end{equation}

\noindent and the average payoff is:

\begin{equation}\label{eq:averagePayoff} 
 u_i(x) =\sum_{j=1}^n x_i^T A_{i j}x_j,
\end{equation}
 
\noindent where $n$ is the number of players with whom player $i$ play the games
and $A_{ij}$ is the payoff matrix among player $i$ and $j$. At each iteration a
player can update his strategy space according to the payoffs gained during the
games, it allocates more probability on the strategies with high payoff, until
an equilibrium is reached, a situation in which it is not possible to obtain
higher payoffs.

To find the Nash equilibrium of the system it is common to use the replicator
dynamic equation \cite{taylor1978evolutionary},

\begin{equation}\label{eq:replicator}
 \dot{x}=[u(e^h)-u(x)] \cdot x^h. \text{} \forall h \in x.
\end{equation}

\noindent This equation allows better than average strategies to increase at
each iteration. It can be used to analyze frequency-dependent selection
processes \cite{nowak2004evolutionary}, furthermore, the fixed points of
equation \ref{eq:replicator} correspond to Nash equilibria
\cite{weibull1997evolutionary}. We used the discrete time version of the
replicator dynamic equation for the experiments of this work.

\begin{equation}\label{eq:repdyn}
  x^h(t+1)=x^h(t)\frac{u(e^h)}{u(x)} \text{ } \forall h \in x(t).
\end{equation}

\noindent The players update their strategies at each time step $t$ considering
the strategic environment in which they are playing.

The complexity of each step of the replicator dynamics is quadratic but there
are more efficient dynamics that can be used, such as the \emph{infection and
immunization} dynamics that has a linear-time/space complexity per step and it
is known to be as accurate as the replicator dynamics \cite{bulo2011graph}.
\section{Dominant Set Clustering}\label{sec:dsclust}%
\emph{Dominant set} is a graph based clustering algorithm that generalizes the
notion of maximal clique from unweighted undirected graphs to edge-weighted
graphs \cite{pavan2007dominant,rota2013game}. With this algorithm it is possible
to extract compact structures from a graph in an efficient way. Furthermore, it
can be used on symmetric and asymmetric similarity graphs and does not require
any parameter. With this framework it is possible to obtain measures of clusters
cohesiveness and to evaluate the strength of participation of a vertex to a
cluster. It models the well-accepted definition of a cluster, which states that
a cluster should have high internal homogeneity and that there should be high
inhomogeneity between the objects in the cluster and those outside
\cite{jain1988algorithms}.

The extraction of compact structures from graphs that reflect these two
conditions, is given by the following quadratic form:

\begin{equation}\label{eq:quadratic} 
  f(x)=x^TAx.
\end{equation}

\noindent Where $A$ is a similarity graph and $x$ is a probability vector, whose
components indicate the participation of each node of the graph to a cluster. In
this context, the clustering task corresponds to the task of finding a vector
$x$ that maximizes $f$ and this can be done with the following program:

\begin{equation}\label{eq:program}
  \begin{multlined} \text{maximize } f(x)\\
  \shoveleft[-1cm]\text{subject to } x \in \Delta. 
  \end{multlined}
\end{equation}

\noindent Where $\Delta$ represents the standard simplex. A (local) solution of
program (\ref{eq:program}) corresponds to a maximally cohesive structure in the
graph \cite{jain1988algorithms}.

The solution of program (\ref{eq:program}) can be found using the discrete time
version of the replicator dynamic equation, computed as follows,

\begin{equation}\label{eq:replicatorDS}
  x(t+1) = x\frac{Ax}{x^T A x },
\end{equation}

\noindent where $x$ represent the strategy space at time $t$.

The clusters are extracted sequentially from the graph using a peel-off strategy
to remove the objects belonging to the extracted cluster, until there are no
more objects to cluster or some predefined criteria are satisfied.

\section{Document Clustering Games}\label{sec:docClustGames}%
In this section we present step by step our approach to document clustering.
First we describe how the documents are represented and how we prepare the data
and structure them using a weighted graph. Then we pass to the preliminary
clustering in order to divide the data points in two disjoint sets of labeled
and unlabeled players. With this information we can initialize the strategy
space of the players and run the dynamics of the system that lead to the final
clustering of the data.
\subsection{Document representation}\label{sec:docRepr}%
The documents of a datasets are processed with a \emph{bag-of-words} (BoW)
model. With this method each document is represented as a vector indexed
according to the frequency of the words in it. To do this it is necessary to
construct the vocabulary of the text collection that is composed by the set of
unique words in the corpus. BoW represents a corpus with a
\textit{document-term matrix}. It consists in a $N \times T$ matrix $M$, where
$N$ is the number of documents in the corpus and $T$ the number of words in the
vocabulary. The words are considered as the features of the documents. Each
element of the matrix $M$ indicates the frequency of a word in a document.

The BoW representation can lead to a high dimensional space, since the
vocabulary size increases as sample size increases. Furthermore, it does not
incorporate semantic information treating homonyms as the same feature. These
problems can result in bad representations of the data and for this reason there
where introduced different approaches to balance the importance of the features
and also to reduce their number, focusing only on the most relevant.

An approach to weigh the importance of a feature is the \emph{term frequency -
inverse document frequency} (tf-idf) method \cite{manning2008introduction}. This
technique takes as input a document-term matrix $M$ and update it with the
following equation,

\begin{equation}\label{eq:tfidf}
  tf \textnormal{-} idf(d,t)=tf(d,t)\cdot log \frac{D}{df(d,t)}, 
\end{equation}

\noindent where $df(d,t)$ is the number of documents that contain word $t$. Then
the vectors are normalized to balance the importance of each feature.

Latent Semantic Analysis (LSA) is a technique used to infer semantic information
\cite{landauer1998introduction} from a \textit{document-term matrix}, reducing
the number of features. Semantic information is obtained projecting the
documents into a \emph{semantic space}, where the relatedness of two terms is
computed considering the number of times they appear in a similar context.
Single Value Decomposition (SVD) is used to create an approximation of the
\textit{document-term matrix} or \emph{tf-idf matrix}. It decomposes a matrix
$M$ in:

\begin{equation}\label{eq:lsa}
  M=U \Sigma V^T,
\end{equation}

\noindent where $\Sigma$ is a diagonal matrix with the same dimensions of $M$
and $U$ and $V$ are two orthogonal matrices. The dimensions of the feature space
is reduced to $k$, taking into account only the first $k$ dimensions of the
matrices in Equation (\ref{eq:lsa}).%

\subsection{Data preparation}\label{sec:dataprep}%
This new representation of the data is used to compute the pairwise similarity
among documents and to construct the proximity matrix $W$, using the cosine
distance as metric,
\begin{equation}\label{eq:cosine}%
  \cos \theta \frac{v_i \cdot v_j}{||v_i|| ||v_j||}
\end{equation}

\noindent where the nominator is the intersection of the words in the vectors
that represent two documents and $||v||$ is the norm of the vectors, which is
calculated as: $ \sqrt{\sum_{i=1}^{n}w_i^2} $.%

\subsection{Graph construction}%
$W$ can be used to represent a text collection as a graph $G$, whose nodes
represent documents and edges are weighted according to the information stored
in $W$. Since, the cosine distance acts as a linear kernel, considering only
information between vectors under the same dimension, it is common to smooth the
data using a kernel function and transforming the proximity matrix $W$ into a
similarity matrix $S$ \cite{shawe2004kernel}. It can also transform a set of
complex and nonlinearly separable patterns into linearly separable patterns
\cite{haykin2004comprehensive}. For this task we used the Gaussian kernel,

\begin{equation}\label{eq:gaussian}
  s_{ij} = exp \left \{ -\frac{w_{ij}^2}{\sigma^2} \right \} 
\end{equation}

\noindent where $w_{ij}$ is the dissimilarity among pattern $i$ and $j$ and
$\sigma$ is a positive real number that determines the kernel width. This
parameter is calculated experimentally, since it is not possible to know in
advance the nature of the data and the clustering separability indices
\cite{peterson2011}. The data representation on the graph can be improved using
graph Laplacian techniques. These techniques are able to decrease the weights of
the edges between different clusters making them more distant. The normalized
graph Laplacian is computed as $L=D^{-1/2} S D^{-1/2}$, where $D$ is the
degree matrix of $S$.

Another technique that can be used to better represent the data is
sparsification, that consists in reducing the number of nodes in the graph,
focusing only on the most important. This refinement is aimed at modeling the
local neighborhood relationships among nodes and can be done with two different
methods, the $\epsilon$-neighborhood technique, which maintains only the edges
which have a value higher than a predetermined threshold, $\epsilon$; and the
$k$-nearest neighbor technique, which maintains only the highest $k$ values. It
results in a similarity matrix that can be used as the adjacency matrix of a
weighted graph $G$.

\begin{figure} \centering
\includegraphics[width=1\textwidth]{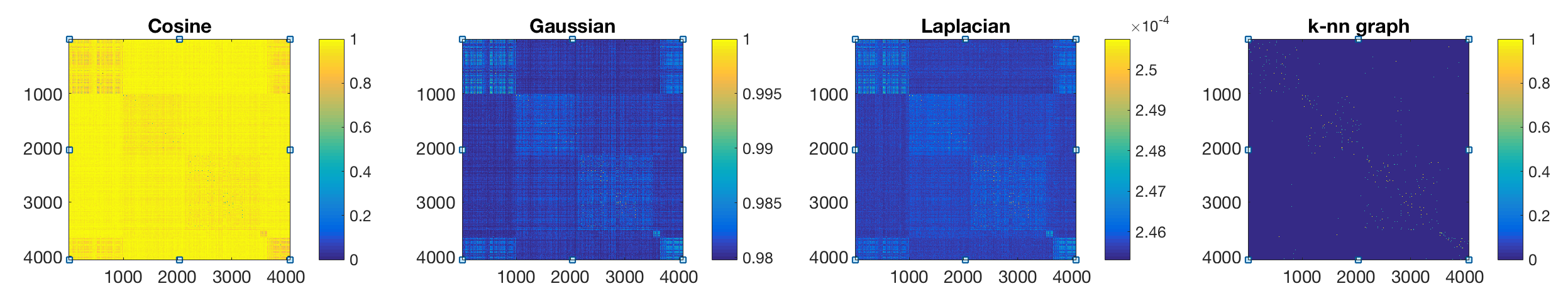}
\caption{Different data representations for a dataset with 5 classes of
different sizes.}\label{fig:wsimGS1S2} \end{figure}

The effect of the processes described above is presented in Fig.
\ref{fig:wsimGS1S2}. Near the main diagonal of the matrices it is possible to
recognize some blocks which represent clusters. The values of those points are
low in the cosine matrix, since it encodes the proximity of the points. Then the
matrix is transformed into a similarity matrix by the Gaussian kernel, in fact,
the values of the points near the main diagonal in this representation are high.
It is possible to note that some noise was removed with the Laplacian matrix.
The points far from the diagonal appear now clearer and the blocks are more
compact. Finally the $k$-nn matrix remove many nodes from the representation,
giving a clear picture of the clusters.

We used the Laplacian matrix $L$ for the experiments with the \emph{dominant
set}, since it requires that the similarity values among the elements of a
cluster are very close to each other. Graph $G$ is used to run the clustering
games, since this framework does not need a dense graph to cluster the data
points.

\subsection{Clustering}\label{sec:clustering}%
We use the \emph{dominant set} algorithm to extract the prototypical elements
of each cluster with two different settings, one in which we give as input the
number of clusters to extract and the other without this information. In the
fist case we extract the first $K$ clusters from a dataset and then run the
document clustering games to cluster the remaining clusters. This situation can
be interpreted as the case in which there are some labeled points in the data
and new points have to be clustered according to this evidence. In the second
case we run \emph{dominant set} recursively to extract small clusters and then
use the document clustering games to cluster the clusters, merging them
according to their similarity. The similarity among two clusters $C_i$ and $C_j$
is computed as:

\begin{equation}\label{eq:simClust}
  sim(C_i,C_j) = \frac{ \sum_{r \in C_i} \sum_{t \in C_j} s_{rt}}{|C_i|+|C_j|}
\end{equation}

We conducted also experiments in which we simulated the streaming data process.
This is done dividing a dataset in random folds and clustering the dataset
iteratively adding a fold at time to measure if the performances of the system
are constant. In this case we used a fold ($8\%$ of the data) as initial
clustering.

\subsection{Strategy space implementation}%
The clustering phase serves as preliminary phase to partition the data into two
disjoint sets, one containing clustered objects and the other unclustered.
Clustered objects supply information to unclustered nodes in the graph. We
initialized the strategy space of the player in these two sets as follows,

\begin{equation}\label{eq:dsstratspace}
x_{i}^h=\begin{cases}
K^{-1}, & \text{if node $i$ is unclustred}.\\ 
1, & \text{if node $i$ is in cluster $h$}, 
\end{cases} 
\end{equation}

\noindent where $K$ is the number of clusters to extract and $K^{-1}$ ensures
that the constraints required by a game theoretic framework are met (see
equation (\ref{eq:simplex})).

\subsection{Clustering games}\label{sec:clustgames}%
We assume that each player $i \in I$ that participates in the games is a
document and that each strategy $s \in S_i$ is a particular cluster. The players
can choose a determined strategy from their strategy space that is initialized
as described in previous section and can be considered as a mixed strategy space
(see Section \ref{sec:GT}). The games are played among two similar players, $i$
and $j$. The payoff matrix among two players $i$ and $j$ is defined as an
identity matrix of rank $K$, $A_{ij}$.

This choice is motivated by the fact that in this context all the players have
the same number of strategies and in the studied contexts the number of clusters
of each dataset is low. In works in which there are many interacting classes it
is possible to use a similarity function to construct the payoff matrix, as
described in \cite{tripodicl2016}.

The best choice for two similar players is to be clustered in the same cluster,
this is imposed with the entry $A_{ij}=1, i=j$. This kind of game is called
\textit{imitation game} because the players try to learn their strategy
observing the choices of their co-players. For this reason the payoff function
of each player is additively separable and is computed as described in Section
\ref{sec:GT}. Specifically, in the case of clustering games there are labeled
and unlabeled players that, as proposed in \cite{erdem2012graph}, can be divided
in two disjoint sets, $I_\mathit{l}$ and $I_\mathit{u}$. We have $K$ disjoint
subsets, $I_\mathit{l}=\{I_{\mathit{l}|1}, ..., I_{\mathit{l}|K} \}$, where each
subset denotes the players that always play their $h$th pure strategy.

Only unlabeled players play the games, because they have to decide their best
strategy (cluster membership). This strategy is selected taking into account the
similarity that a player share with other players and the choices of these
players. Labeled players act as bias over the choices of unlabeled players
because they always play a defined strategy and unlabeled players influence each
other. The players adapt to the strategic environment, gradually adjusting their
preferences over strategies \cite{sandholm2010population}. Once equilibrium is
reached, the cluster of each player $i$, corresponds to the strategy, with the
highest value.

The payoffs of the games are calculated with equations \ref{eq:singlePayoff} and
\ref{eq:averagePayoff}, which in this case, with labeled and unlabeled players,
can be defined as, 

\begin{equation}\label{eq:singleTransuction} u_i(e_i^h)= \sum_{j \in I_\mathit{u}} (g_{ij} A_{ij} x_j)^h + \sum_{h=1}^K \sum_{j \in I_{\mathit{l}|h}} (g_{ij} A_{ij})^h \end{equation}

\noindent and,

\begin{equation}\label{eq:averageTransuction} %
  u_i(x)= \sum_{j \in I_\mathit{u}} x_i^T g_{ij} A_{ij} x_j + \sum_{k=1}^K \sum_{j \in I_{\mathit{l}|h}} x_i^T (g_{ij} A_{ij} )^h. 
\end{equation}

\noindent where the first part of the equations calculates the payoffs that each
player obtains from unclustered players and the second part computes the
payoffs obtained from labeled players. The Nash equilibria of the system are
calculated the replicator dynamics equation \ref{eq:repdyn}.

\section{Experimental Setup}%
The performances of the systems are measured using the accuracy (AC) and the
normalized mutual information (NMI). AC is calculated with the following
equation,

\begin{equation}\label{eq:acc}
AC=\frac{\sum_{i=1}^n{\delta(\alpha_i,map(l_i))}}{n},
\end{equation}

\noindent where $n$ denotes the total number of documents in the dataset and
$\delta(x,y)$ is equal to $1$ if $x$ and $y$ are clustered in the same cluster.
the function $map(L_i)$ maps each cluster label $l_i$ to the equivalent label in
the benchmark, aligning the labeling provided by the benchmark and those
obtained with our clustering algorithm. It is done using the Kuhn-Munkres
algorithm \cite{lovasz1986matching}. The NMI was introduced by Strehl and Ghosh
\cite{strehl2003cluster} and indicates the level of agreement between the
clustering $C$ provided by the ground truth and the clustering $C'$ produced by
a clustering algorithm. This measure takes into account also the partitioning
similarities of the two clustering and not just the number of correctly
clustered objects. The mutual information (MI) between the two clusterings is
computed with the following equation,

\begin{equation}\label{eq:mi}
MI(C,C')=\sum_{c_i \in C, c_j' \in C'}p(c_i,c_j') \cdot log_2 \frac{p(c_i,c_j')}{p(c_i) \cdot p(c_j')},
\end{equation}

\noindent where $p(c_i)$ and $p(c_i')$ are the probabilities that a document
belongs to cluster $c_i$ and $c_i'$, respectively; $p(c_i,c_i')$ is the
probability that the selected document belongs to $c_i$ as well as $c_i'$ at the
same time. The MI information is then normalized with the following equation,

\begin{equation}\label{eq:nmi} 
  NMI(C,C')=\frac{MI(C,C')}{max(H(C),H(C'))}
\end{equation} 

\noindent where $H(C)$ and $H(C')$ are the entropies of $C$ and $C'$,
respectively, This measure ranges from 0 to 1. It is equal to $1$ when the two
clustering are identical and it is equal to $0$ if the two sets are independent.
We run each experiment 50 times and present the mean results with standard
deviation ($\pm$). 

We evaluated our model on the same
datasets\footnote{\url{http://www.shi-zhong.com/software/docdata.zip} .} used in
\cite{zhong2005generative}. In that work it has been conducted an extensive
comparison of different document clustering algorithms. The description of these
datasets is shown in Table \ref{tab:datasets}. The authors used 13 datasets
(described in Table \ref{tab:datasets}). The datasets have different sizes
($n_d$), from 204 documents (tr23) to 8580 (\emph{sports}). The number of
classes ($K$) is also different and ranges from 3 to 10. Another important
feature of the datasets is the size of the vocabulary ($n_w$) of each dataset
that ranges from 5832 (\emph{tr23}) to 41681 (\emph{classic}) and is function of
the number of documents in the dataset, their size and the number of different
topics in it, that can be considered as clusters. The datasets are also
described with $n_c$ and $Balance$. $n_c$ indicates the average number of
documents per cluster and $Balance$ is the ratio among the size of the smallest
cluster and that of the largest.

\begin{table}
\begin{center}
\begin{tabular}{| l | l l l l l |} \hline
Data & $n_d$ & $n_v$ & K & $n_c$ & Balance  \\ \hline
NG17-19 & 2998 & 15810 & 3 & 999 & 0.998 \\ 
classic & 7094 & 41681 & 4 & 1774 & 0.323 \\
k1b & 2340 & 21819 & 6 & 390 & 0.043 \\
hitech & 2301 & 10800 & 6 & 384 & 0.192 \\
reviews & 4069 & 18483 & 5 & 814 & 0.098 \\
sports & 8580 & 14870 & 7 & 1226 & 0.036 \\
la1 & 3204 & 31472 & 6 & 534 & 0.290 \\
la12 & 6279 & 31472 & 6 & 1047 & 0.282 \\
la2 & 3075 & 31472 & 6 & 513 & 0.274 \\
tr11 & 414 & 6424 & 9 & 46 & 0.046 \\
tr23 & 204 & 5831 & 6 & 34 & 0.066 \\
tr41 & 878 & 7453 & 10 & 88 & 0.037 \\
tr45 & 690 & 8261 & 10 & 69 & 0.088 \\\hline
\end{tabular}
\end{center}
\caption{Datasets description}
\label{tab:datasets}
\end{table}

\subsection{Basic Experiments}%
We present in this Section an experiment in which all the features of each
dataset are used, constructing the graphs as described in Section
\ref{sec:docClustGames}. We first used \emph{dominant set} to extract the
prototypical elements of each cluster and then we applied our approach to
cluster the remaining data points.

The results of this series of experiments are presented in Table
\ref{tab:risDSentireFeat1}. They can be used as point of comparison for our next
experiments, in which we used different settings. From the analysis
of the table it is not possible to find a stable pattern. The results range from
NMI $.27$ on the \textit{hitech}, to NMI $.67$ on \textit{k1b}. The reason of
this instability is due to the representation of the datasets that in some cases
is not appropriate to describe the data.

\begin{figure} \centering
\includegraphics[trim={2cm 1cm 2cm 0},clip,width=1\textwidth]{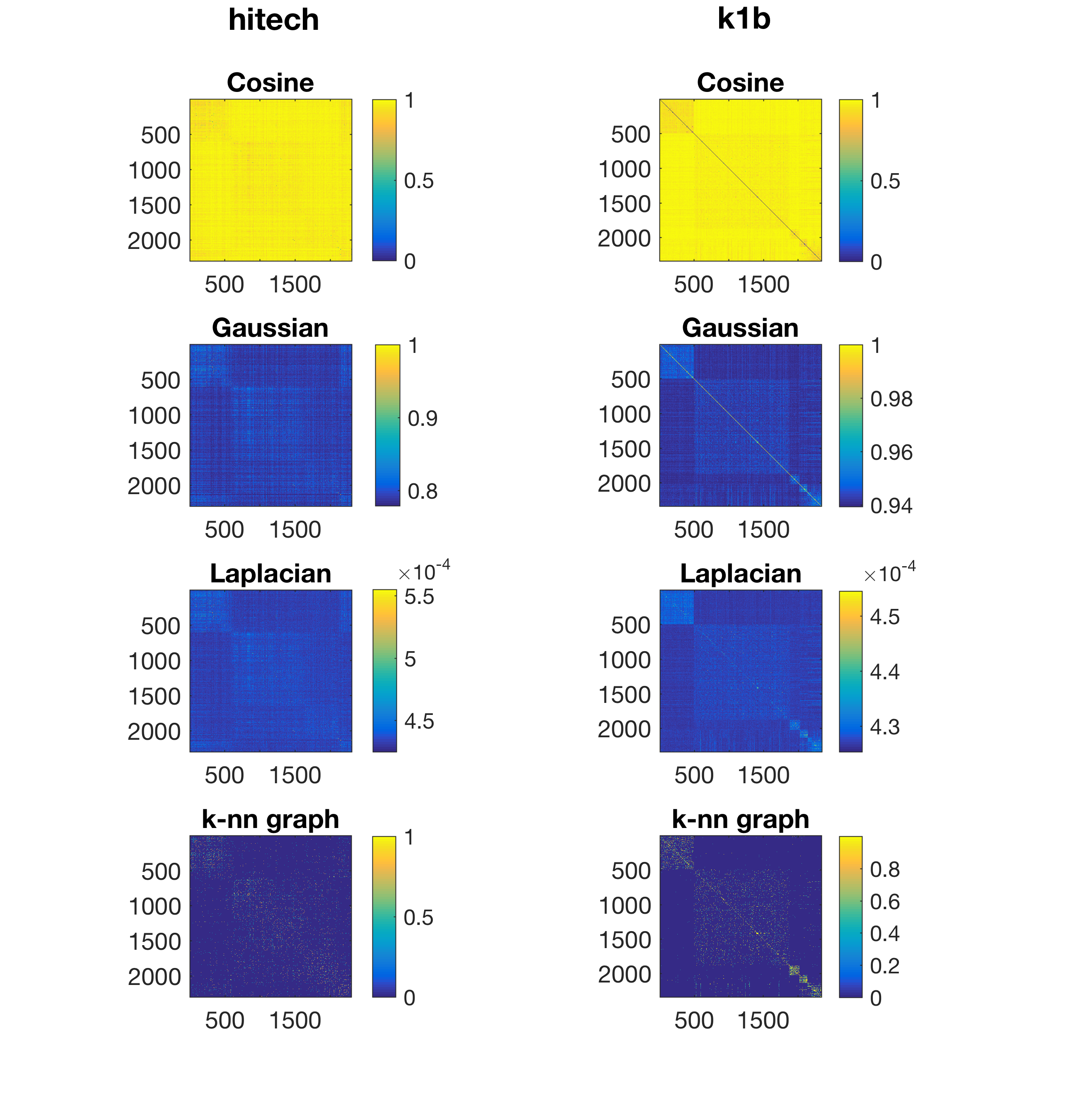}
\caption{Different representations for the datasets \textit{hitech} and \textit{k1b}.}\label{fig:k1bhitech}
\end{figure}

An example of the graphical representation of the two datasets mentioned above
is presented in Fig. \ref{fig:k1bhitech}, where the similarity matrices
constructed for \textit{k1b} and \textit{hitech} are shown. We can see that the
representation of \textit{hitech} does not show a clear structure near the main
diagonal, to the contrary, it is possible to recognize a block structures on the
graphs representing \textit{k1b}.

\begin{table}
\begin{center}
\resizebox{\textwidth}{!}{%
\begin{tabular}{| l | l | l | l | l | l | l | l | l | l | l | l | l | l | l | l |  |} \hline
&NG17-19 & classic & k1b & hitech & review & sports & la1 & la12 & la2 & tr11 & tr23 & tr41 & tr45 \\ \hline
AC& $.56 \pm .0$  & $.66 \pm .07$ & $.82 \pm .0$ & $.44 \pm .0$ & $.81 \pm .0$ & $.69 \pm .0$ & $.49 \pm .04$ & $.57 \pm .02$ & $.54 \pm .0$ & $.68 \pm .02$ & $.44 \pm .01$ & $.64 \pm .07$ & $.64 \pm .02$ \\
NMI& $.42 \pm .0$ & $.56 \pm .22$ & $.66 \pm .0$   & $.27 \pm .0$ & $.59 \pm .0$ & $.62 \pm .0$  & $.45 \pm .04$  & $.46 \pm .01$ & $.46 \pm .01$ & $.63 \pm .02$ & $.38 \pm .0$ & $.53 \pm .06$ & $.59 \pm .01$ \\ \hline
\end{tabular} }
\end{center}
\caption{Results as AC and NMI, with
the entire feature space.}
\label{tab:risDSentireFeat1}
\end{table}

\subsection{Experiments with Feature Selection}%
In this section we present an experiment in which we conducted feature selection
to see if it is possible to reduce the noise introduced by determined features.
To do this, we decided to apply to the corpora a basic frequency selection
heuristic that eliminates the features that occur more (or less) often than a
determined thresholds. In this study were kept only the words that occur more
than once.

This basic reduction leads to a more compact feature space, which is easier to
handle. Words that appear very few times in the corpus can be special characters
or miss-spelled words and for this reason can be eliminated. The number of
features of the reduced datasets are shown in Table \ref{tab:freqSel}. From the
table, we can see that the reduction is significant for $5$ of the datasets
used, with a reduction of $82\%$ for \textit{classic}. The datasets that are not
listed in the table were not affected by this process.

In Table \ref{tab:risBasicFeatureSelection} we present the results obtained
employing feature selection. This technique can be considered a good choice to
reduce the size of the datasets and the computational cost, but in this case
does not seem to have a big impact on the performances of the algorithm. In
fact, the improvements in the performance of the algorithm are not substantial.
There is an improvement of $1\%$, in terms of $NMI$, in four datasets over five
and in one case we obtained lower results. This could be due to the fact that we
do not know exactly what features have been removed, because this information is
not provided with the datasets. It is possible that the reduction has removed
some important (discriminative) word, compromising the representation of the
data and the computation of the similarities. Also for this reason we did not
use any other frequency selection technique.

\begin{table}
\resizebox{0.40\textwidth}{!}{\begin{minipage}[a]{0.44\textwidth}
\begin{tabular}{| l | c | c | c | c | c | c | } \hline
& classic & k1b & la1 & la12 & la2  \\ \hline
pre& 41681 & 21819 & 31472 & 31472 & 31472  \\ \hline
post& 7616 & 10411 &  13195 & 17741 & 12432  \\ \hline
\% & 0.82 & 0.52 & 0.58 & 0.44 & 0.6 \\ \hline
\end{tabular}\vspace{12pt}
\caption{Number of features for each dataset before and after feature 
selection.}
\end{minipage}\label{tab:freqSel} }
\hfill
\resizebox{0.5\textwidth}{!}{\begin{minipage}[a]{0.57\textwidth}
\begin{tabular}{| l | c | c | c | c | c | c | } \hline
 & classic & k1b & la1 & la12 & la2  \\ \hline
AC  & $.67 \pm .0$ & $.79 \pm .0$ & $.56 \pm .11$ & $.56 \pm .03$ & $.57 \pm .0$  \\
NMI   & $.57 \pm .0$ & $.67 \pm .0$ & $.47 \pm .12$ & $.44 \pm .01$ & $.47 \pm .0$  \\ \hline 
\end{tabular}\vspace{12pt}
\caption{Mean results as AC and NMI, with frequency 
selection.}
\end{minipage}\label{tab:risBasicFeatureSelection}}
\end{table}

\subsection{Experiments with LSA}%
In this Section we used LSA (see Section \ref{sec:docRepr}) to reduce the number
of features that describe the data. The evaluation was conducted using different
numbers of features to describe each dataset, ranging from 10 to 400. This
operation is required because there is no agreement on the correct number of
features to extract for a determined dataset, for this reason this value has to
be calculate experimentally.

The results of this evaluation are shown in two different tables, Table
\ref{tab:LSAcomptsMI} indicates the results as NMI and Table
\ref{tab:LSAcomponentsAcc} indicates the results as AC for each dataset and
number of features. The performances of the algorithm measured as NMI are
similar on average (excluding the case of $n_v$ with 10 features), but there is
no agreement on different datasets. In fact, different data representations
affect heavily the performances on datasets such as \emph{NG17-19}, where the
performances ranges from $.27$ to $.46$. This phenomenon is due to the fact that
each dataset has different characteristics, as shown in Table \ref{tab:datasets}
and that their representation require an appropriate semantic space. With $n_v =
250$ we obtained the higher results on average, both in terms of NMI and AC.

The results with the representation provided by LSA show how this technique is
effective in terms of performances. In fact, it is possible to achieve higher
results than using the entire feature space or with the frequency selection
technique. The improvements are substantial and in many cases are $10\%$ higher.
Furthermore, with this new representation it is easier to handle the data.

\begin{table}
\resizebox{0.475\textwidth}{!}{\begin{minipage}[a]{0.55\textwidth}
\begin{tabular}{| l | c c c c c c c c c |} \hline
Data$\backslash n_v$& 10 & 50 & 100 & 150 & 200 & 250 & 300 & 350 & 400  \\ \hline
NG17-19 & $.27$ & $.37$ & $\mathbf{.46}$ & $.26$ & $.35$ & $.37$ & $.36$ & $.37$ & $.37$  \\
classic & $.53$ & $.63$ & $.71$ & $.73$ & $\mathbf{.76}$ & $.74$ & $.72$ & $.72$ & $.69$ \\
k1b &  $\mathbf{.68}$ & $.61$ & $.58$ & $.62$ & $.63$ & $.63$ & $.62$ & $.61$ & $.62$ \\
hitech & $\mathbf{.29}$ & $.28$ & $.25$ & $.26$ & $.28$ & $.27$ & $.27$ & $.26$ & $.26$ \\
reviews & $\mathbf{.60}$ & $.59$ & $.59$ & $.59$ & $.59$ & $.59$ & $.58$ & $.58$ & $.58$   \\
sports & $.62$ & $.63$ & $\mathbf{.69}$ & $.67$ & $.66$ & $.66$ & $.66$ & $.64$ & $.62$ \\
la1 & $.49$ & $.53$ & $.58$ & $.58$ & $.58$ & $.57$ & $\mathbf{.59}$ & $.57$ & $\mathbf{.59}$  \\
la12 & $.48$ & $.52$ & $.52$ & $.52$ & $.53$ & $\mathbf{.56}$ & $.54$ & $.55$ & $.54$ \\
la2 &  $.53$ & $.56$ & $.58$ & $.58$ & $.58$ & $.58$ & $\mathbf{.59}$ & $.58$ & $.58$\\
tr11 & $.69$ & $.65$ & $.67$ & $.68$ & $\mathbf{.71}$ & $.70$ & $.70$ & $.69$ & $.70$ \\
tr23 & $.42$ & $\mathbf{.48}$ & $.41$ & $.39$ & $.41$ & $.40$ & $.41$ & $.40$ & $.41$  \\
tr41 & $.65$ & $.75$ & $.72$ & $.69$ & $.71$ & $.74$ & $\mathbf{.76}$ & $.69$ & $.75$ \\
tr45 & $.65$ & $\mathbf{.70}$ & $.67$ & $.69$ & $.69$ & $.68$ & $.68$ & $.67$ & $.69$ \\ \hline
avg. & $.53$  &  $.56$  &  $\mathbf{.57}$  &  $.56$  &  $\mathbf{.57}$  &  $\mathbf{.57}$  &  $\mathbf{.57}$  &  $.56$  &  $\mathbf{.57}$ 
\\ \hline
\end{tabular}\vspace{12pt}
\caption{NMI results for all the datasets.
Each column indicates the results obtained with a reduced version of the feature space using LSA.}
\label{tab:LSAcomptsMI}
\end{minipage}}
\hfill
\resizebox{0.475\textwidth}{!}{\begin{minipage}[a]{0.55\textwidth}
\begin{tabular}{| l | c c c c c c c c c  |} \hline
Data$\backslash n_v$& 10 & 50 & 100 & 150 & 200 & 250 & 300 & 350 & 400  \\ \hline
NG17-19 & $.61$ & $\mathbf{.63}$ & $.56$ & $.57$ & $.51$ & $.51$ & $.51$ & $.51$ & $.51$  \\
classic & $.64$ & $.76$ & $.87$ & $.88$ & $\mathbf{.91}$ & $.88$ & $.85$ & $.84$ & $.80$  \\
k1b & $.72$ & $.55$ & $.58$ & $.73$ & $\mathbf{.75}$ & $\mathbf{.75}$ & $.73$ & $.70$ & $.73$ \\
hitech & $\mathbf{.48}$ & $.36$ & $.42$ & $.41$ & $.47$ & $.46$ & $.41$ & $.43$ & $.42$  \\
reviews &  $\mathbf{.73}$ & $.72$ & $.69$ & $.69$ & $.69$ & $.71$ & $.71$ & $.71$ & $.71$ \\
sports & $.62$ & $.61$ & $\mathbf{.71}$ & $.69$ & $.68$ & $.68$ & $.68$ & $.68$ & $.61$ \\
la1 & $.59$ & $.64$ & $.72$ & $.70$ & $\mathbf{.73}$ & $.72$ & $\mathbf{.73}$ & $.72$ & $\mathbf{.73}$ \\
la12 & $.63$ & $.63$ & $.62$ & $.62$ & $.63$ & $\mathbf{.67}$ & $.64$ & $\mathbf{.67}$ & $.65$ \\
la2 & $\mathbf{.69}$ & $.66$ & $.60$ & $.60$ & $.61$ & $.60$ & $.65$ & $.60$ & $.60$ \\
tr11 & $.69$ & $.66$ & $.69$ & $.70$ & $\mathbf{.72}$ & $.71$ & $.71$ & $.71$ & $.71$ \\
tr23 &  $.44$ & $\mathbf{.51}$ & $.43$ & $.42$ & $.43$ & $.43$ & $.43$ & $.43$ & $.43$ \\
tr41 & $.60$ & $.76$ & $.68$ & $.68$ & $.65$ & $.75$ & $.\mathbf{77}$ & $.67$ & $.\mathbf{77}$ \\
tr45 & $.57$ & $\mathbf{.69}$ & $.66$ & $.68$ & $.67$ & $.67$ & $.67$ & $.67$ & $.67$ \\  \hline
avg. & .$62$  &  $.63$  &  $.63$  &  $ .64$  &  $.65$  &  $\mathbf{.66}$  &  $.65$  &  $.64$  &  $.64$
\\ \hline
\end{tabular}\vspace{12pt}
\caption{AC results for all the datasets. Each column
indicates the results obtained with a reduced version of the feature space using
LSA.}\label{tab:LSAcomponentsAcc}
\end{minipage}}
\end{table}

\subsection{Comparison with State-of-the-art algorithms}\label{sec:compsota}%
The results of the evaluation of the document clustering games are shown in
Table \ref{tab:noClassNMI} and \ref{tab:noClassAC} (third column, DCG). We
compared the best results obtained with the document clustering games approach
and the best results indicated in \cite{zhong2005generative} and in
\cite{pompili2014two}. In the first article it was conducted an extensive
evaluation of different generative and discriminative models, specifically
tailored for document clustering and two graph-based approaches, CLUTO and a
bipartite spectral co-clustering method. In this evaluation the results are
reported as NMI and graphical approaches obtained better performances than
generative. In the second article were evaluated different NMF approaches to
document clustering, on the same datasets, here the results are reported as AC.

From Table \ref{tab:noClassNMI} it is possible to see that the results of the
document clustering games are higher than those of state-of-the-art algorithms
on ten datasets out of thirteen. On the remaining three datasets we obtained the
same results on two datasets and a lower result in one. On \emph{classic},
\emph{tr23} and tr26 the improvement of our approach is substantial, with
results higher than $5\%$. Form Table \ref{tab:noClassAC} we can see that our
approach performs substantially better that NMF on all the datasets.

\begin{table}
\resizebox{0.455\textwidth}{!}{\begin{minipage}[a]{0.465\textwidth}
\begin{tabular}{| l | l l l |} \hline
Data & $DCG_{noK}$ & $DCG$ & $Best$  \\ \hline
NG17-19 & $.39 \pm .0 $ & $\textbf{.46} \pm .0$ & $\textbf{.46} \pm .01$ \\
classic & $.71 \pm .0 $ & $\textbf{.76} \pm .0$ & $.71 \pm .06$ \\
k1b & $\textbf{.73}\pm .02 $ & $.68 \pm .02$  & $.67 \pm .04$ \\
hitech & $\textbf{.35} \pm .01 $ & $.29 \pm .02$ & $.33 \pm .01$ \\
reviews & $.57 \pm .01 $ & $\textbf{.60} \pm .01$ & $.56 \pm .09$ \\
sports & $.67 \pm .0 $ & $\textbf{.69} \pm .0$ & $.67 \pm .01$ \\
la1 & $.53 \pm .0 $ & $\textbf{.59} \pm .0$ & $.58 \pm .02$ \\
la12 & $.52 \pm .0 $ & $\textbf{.56} \pm .0$ & $\textbf{.56} \pm .01$ \\
la2 & $.53 \pm .0 $ & $\textbf{.59} \pm .0$ & $.56 \pm .01$ \\
tr11 & $\textbf{.72}\pm .0$ & $.71 \pm .0$ & $.68 \pm .02$ \\
tr23  & $\textbf{.57}\pm .02$ & $.48 \pm .03$ & $.43 \pm .02$ \\
tr41  & $.70 \pm .01 $ & $\textbf{.76} \pm .06$ & $.69 \pm .02$ \\
tr45  & $\textbf{.70} \pm .02$ & $\textbf{.70} \pm .03$ & $.68 \pm .05$ \\ \hline
\end{tabular}\vspace{12pt}
\caption{Results as NMI of generative models and graph partitioning algorithm (\emph{Best}) compared to our approach with and without $k$.}\label{tab:noClassNMI}
\end{minipage}}
\hfill
\resizebox{0.455\textwidth}{!}{\begin{minipage}[a]{0.465\textwidth}
\begin{tabular}{| l | l l l |} \hline
Data & $DCG_{noK}$ & $DCG$ & $Best$  \\ \hline
NG17-19 & $ .59 \pm .0$ & $\textbf{.63} \pm .0$ & -  \\
classic & $ .80 \pm .0$ & $\textbf{.91} \pm .0$ & $.59 \pm .07$  \\
k1b & $\textbf{.86} \pm .02$ & $.75 \pm .03$ & $.79 \pm .0 $  \\
hitech & $\textbf{.52} \pm .01$ & $.48 \pm .02$ & $.48 \pm .04$  \\
reviews & $ .64 \pm .01$ & $\textbf{.73} \pm .01$ & $.69 \pm .07$  \\
sports & $\textbf{.78} \pm .0$ & $.71 \pm .0$ & $.50 \pm .07$  \\
la1 & $ .63 \pm .0$ & $\textbf{.73} \pm .0$ & $.66 \pm .0$  \\
la12 & $ .59 \pm .0$ & $\textbf{.67} \pm .0$ & -  \\
la2 & $ .55 \pm .0$ & $\textbf{.69} \pm .0$ & $.53 \pm .0$  \\
tr11 & $\textbf{.74} \pm .0$ & $.72 \pm .0$ & $.53 \pm .05$  \\
tr23  & $\textbf{.52} \pm .02$  & $.51 \pm .05$ & $.43 \pm .06$  \\
tr41  & $.75  \pm .01 $ & $\textbf{.77} \pm .08$ & $.53 \pm .06$  \\
tr45  & $\textbf{.71} \pm .01$ & $.69 \pm .04$ & $.54 \pm .06$ \\ \hline
\end{tabular}\vspace{12pt}
\caption{Results as AC of nonnegative matrix factorization algorithms (\emph{Best}) compared to our approach with and without $k$.}\label{tab:noClassAC}
\end{minipage}}
\end{table}

\subsection{Experiments with no Cluster Number}\label{sec:nok}
In this section we present the experiments conducted with our system in a
context in which the number of clusters to extract from the dataset is not used.
It has been tested the ability of \emph{dominant set} to find natural clusters
and the performances that can be obtained in this context by the document
clustering games. We first run \emph{dominant set} to discover many small
clusters, setting the parameter of the gaussian kernel with a small value
($\sigma = 0.1$), then these clusters are re-clustered as described in Section
\ref{sec:clustering} constructing a graph that encodes their pairwise similarity
(see equation \ref{eq:simClust}).

\begin{figure} \centering \includegraphics[trim={2cm 1cm 2cm
0},clip,width=0.7\textwidth]{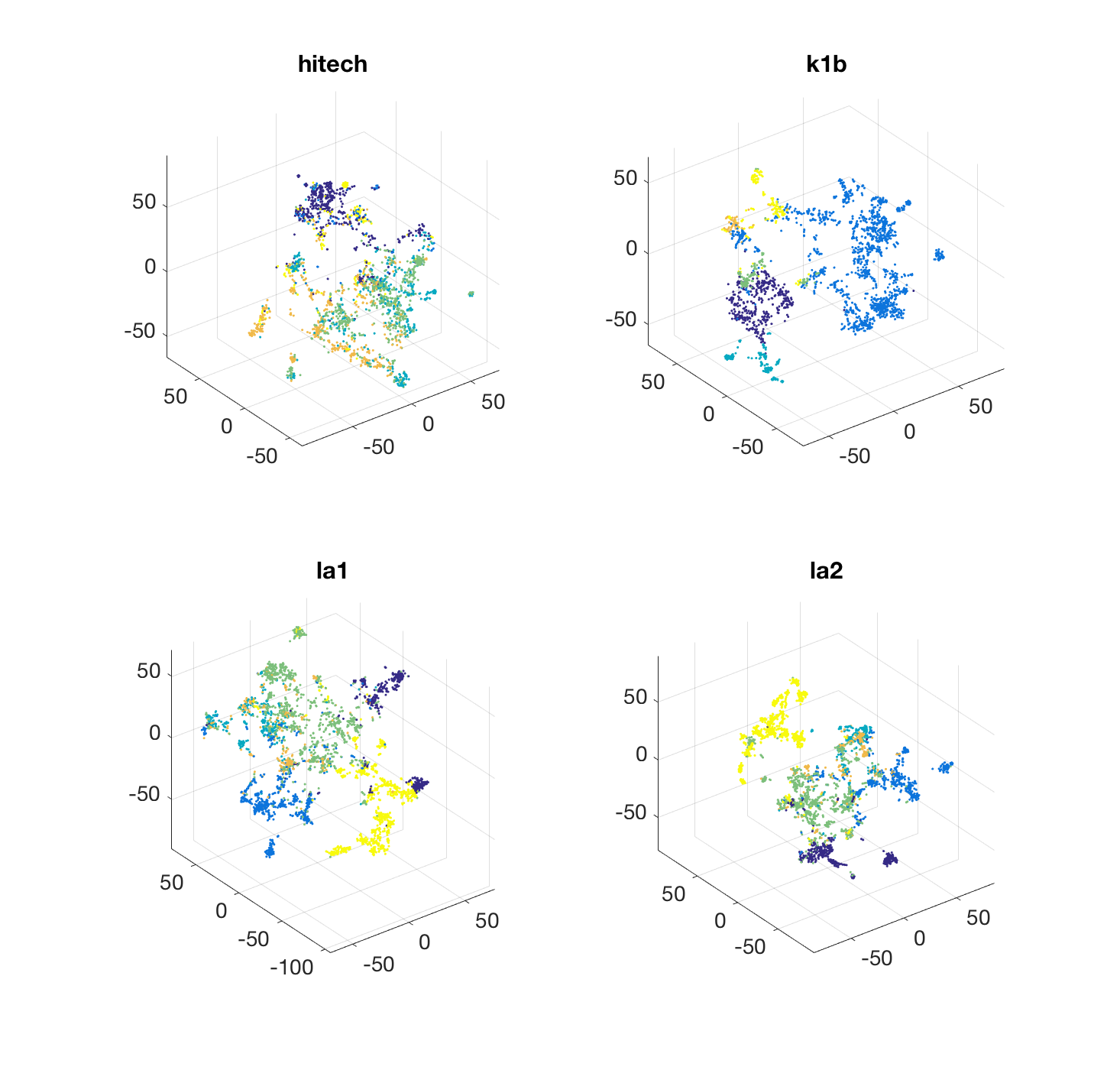}
\caption{Representation of the datasets \textit{hitech}, \textit{k1b},
\textit{la1} and \textit{la2}.}\label{fig:gscatter3} \end{figure}

The evaluation of this model was conducted on the same datasets used in previous
experiments and the results are shown in Table \ref{tab:noClassNMI} and
\ref{tab:noClassAC} (second column, $DCG_{noK}$). From these tables we can see
that this new formulation of the clustering games performs well in many
datasets. In fact, in datasets such as \emph{k1b}, \emph{hitech}, \emph{tr11}
and \emph{tr23} it has results higher than those obtained in previous
experiments. This can be explained by the fact that with this formulation the
number of clustered points used by our framework is higher that in the previous
experiments. Furthermore, this new technique is able to extract clusters of any
shape. In fact, as we can see in Fig. \ref{fig:gscatter3}, datasets such as
\emph{la1} and \emph{la2} present a more compact cluster structure, whereas in
datasets such as \emph{k1b} and \emph{hitech} the clusters structure is
loose\footnote{The dataset have been visualized using t-SNE to reduce the
features to 3d.}.

The performances of the system can be improved with this setting when it
is able to extract the exact number of natural clusters from the graph. To the
contrary, when it is not able to predict this number, the performances decrease
drastically. This phenomenon can explain why this approach performs poorly in
some datasets. In fact, in datasets such as, \emph{NG17-19}, \emph{la1},
\emph{la12} and \emph{l2} the system performs poorly compared to our previous
experiments. In many cases this happens because during the clustering phase we
extract more clusters than expected. The results as NMI of our system are higher
than those of related algorithms on $8$ over $13$ datasets, even if $k$ is not
given as input. Also the results as AC are good, in fact on $9$ datasets over
$11$ we obtained better performances.

\subsection{Experiments on streaming data}\label{sec:expStream}%
In this section we present the evaluation of our approach on streaming datasets.
For this task we used the same datasets used in previous experiments but this
time we divided each of them in 12 random folds. In this way we simulated the
data streaming process, clustering the data iterativelly. We performed the
experiments 15 times to not bias our test sets. For each experiment we selected
a random fold as initial clustering and performed 11 runs of the algorithm, each
time including a new fold in the test set. Previous clusterings are used to
drive the choices of new data points to specific clusters, making the final
clustering coherent.

\begin{table}
  \resizebox{\textwidth}{!}{%
\begin{tabular}{| l | c | c | c | c | c | c | c | c | c | c | c |} \hline
Data & 2 & 3 & 4 & 5 & 6 & 7 & 8 & 9 & 10 & 11 & 12  \\ \hline
ng17-19 & $.57 \pm .07$ & $.55 \pm .05$ & $.55 \pm .04$ & $.55 \pm .03$ & $.55 \pm .03$ & $.55 \pm .03$ & $.55 \pm .03$ & $.55 \pm .03$ & $.55 \pm .03$ & $.55 \pm .03$ & $.55 \pm .03$\\ 
classic & $.81 \pm .02$ & $.81 \pm .02$ & $.81 \pm .02$ & $.81 \pm .01$ & $.81 \pm .01$ & $.81 \pm .01$ & $.81 \pm .01$ & $.81 \pm .01$ & $.81 \pm .01$ & $.81 \pm .01$ & $.81 \pm .01$\\ 
k1b & $.85 \pm .04$ & $.83 \pm .03$ & $.83 \pm .02$ & $.83 \pm .02$ & $.83 \pm .02$ & $.83 \pm .01$ & $.83 \pm .01$ & $.83 \pm .01$ & $.83 \pm .02$ & $.83 \pm .01$ & $.83 \pm .01$\\ 
hitech & $.38 \pm .04$ & $.34 \pm .04$ & $.34 \pm .03$ & $.33 \pm .03$ & $.33 \pm .02$ & $.32 \pm .02$ & $.32 \pm .02$ & $.32 \pm .02$ & $.32 \pm .02$ & $.32 \pm .02$ & $.32 \pm .02$\\ 
reviews & $.77 \pm .03$ & $.75 \pm .02$ & $.75 \pm .02$ & $.74 \pm .02$ & $.74 \pm .01$ & $.74 \pm .02$ & $.74 \pm .02$ & $.74 \pm .01$ & $.74 \pm .02$ & $.74 \pm .01$ & $.74 \pm .01$\\ 
sports & $.86 \pm .02$ & $.85 \pm .02$ & $.84 \pm .02$ & $.84 \pm .01$ & $.84 \pm .01$ & $.83 \pm .01$ & $.83 \pm .01$ & $.83 \pm .01$ & $.83 \pm .01$ & $.83 \pm .01$ & $.83 \pm .01$\\ 
la1 & $.65 \pm .05$ & $.63 \pm .04$ & $.63 \pm .04$ & $.63 \pm .03$ & $.64 \pm .02$ & $.64 \pm .02$ & $.63 \pm .02$ & $.63 \pm .02$ & $.63 \pm .02$ & $.63 \pm .02$ & $.63 \pm .02$\\ 
la12 & $.68 \pm .03$ & $.67 \pm .02$ & $.66 \pm .01$ & $.67 \pm .01$ & $.66 \pm .01$ & $.66 \pm .01$ & $.66 \pm .01$ & $.66 \pm .01$ & $.66 \pm .01$ & $.66 \pm .01$ & $.66 \pm .01$\\ 
la2 & $.68 \pm .03$ & $.67 \pm .02$ & $.67 \pm .02$ & $.67 \pm .02$ & $.66 \pm .02$ & $.66 \pm .01$ & $.66 \pm .01$ & $.67 \pm .01$ & $.67 \pm .01$ & $.67 \pm .01$ & $.67 \pm .02$\\ 
tr11 & $.69 \pm 10$ & $.64 \pm .09$ & $.61 \pm 10$ & $.58 \pm .08$ & $.56 \pm .08$ & $.56 \pm .07$ & $.55 \pm .07$ & $.54 \pm .07$ & $.54 \pm .07$ & $.54 \pm .07$ & $.54 \pm .07$\\ 
tr23 & $.66 \pm 11$ & $.57 \pm 10$ & $.52 \pm .08$ & $.50 \pm .09$ & $.50 \pm .08$ & $.49 \pm .08$ & $.48 \pm .09$ & $.48 \pm .09$ & $.47 \pm .08$ & $.46 \pm .08$ & $.45 \pm .08$\\ 
tr41 & $.86 \pm .05$ & $.84 \pm .05$ & $.83 \pm .04$ & $.83 \pm .04$ & $.83 \pm .03$ & $.82 \pm .03$ & $.82 \pm .03$ & $.82 \pm .03$ & $.82 \pm .03$ & $.82 \pm .03$ & $.81 \pm .03$\\ 
tr45 & $.79 \pm .04$ & $.76 \pm .04$ & $.76 \pm .04$ & $.75 \pm .04$ & $.74 \pm .04$ & $.74 \pm .04$ & $.73 \pm .04$ & $.73 \pm .03$ & $.73 \pm .03$ & $.73 \pm .04$ & $.73 \pm .04$
\\ \hline
\end{tabular}}
\vspace{12pt}
\caption{Results as NMI for all the datasets.
Each column indicates the results obtained including the corresponding fold in the test set.}
\label{tab:streamingMI}

\end{table}

\begin{figure} \centering \includegraphics[trim={10cm 4cm 6cm
4},clip,width=1\textwidth]{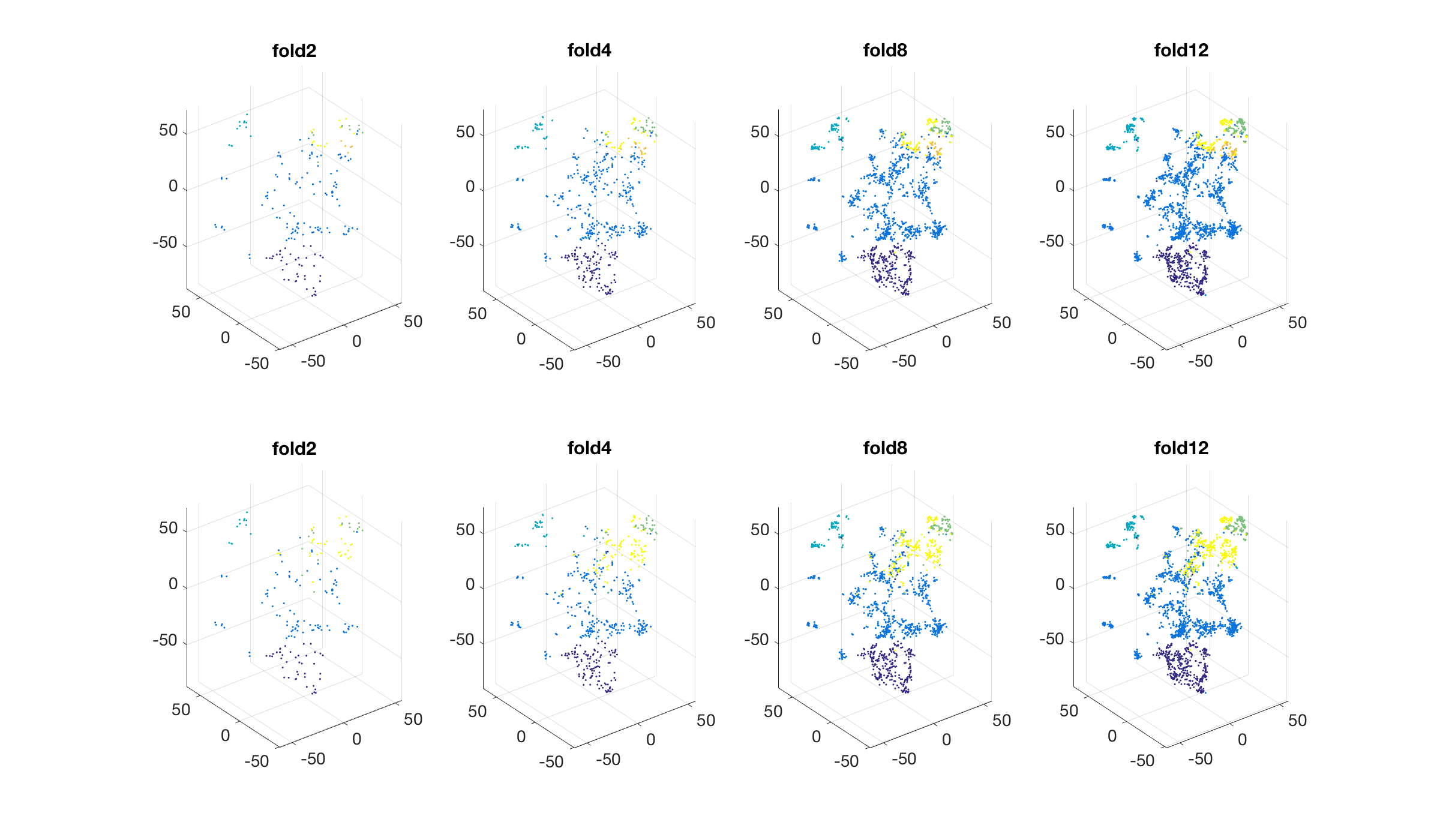}
\caption{Visualizations of the results on \emph{k1b} on different folds. The
upper row shows the ground truth and the lower row shows the results of our
approach.}\label{fig:streaming} \end{figure}

\begin{table}
  \resizebox{\textwidth}{!}{%
\begin{tabular}{| l | c | c | c | c | c | c | c | c | c | c | c | c |} \hline
Data & 2 & 3 & 4 & 5 & 6 & 7 & 8 & 9 & 10 & 11 & 12  \\ \hline
ng17-19 & $.85 \pm .03$ & $.84 \pm .03$ & $.84 \pm .02$ & $.84 \pm .01$ & $.84 \pm .02$ & $.84 \pm .01$ & $.84 \pm .01$ & $.84 \pm .01$ & $.84 \pm .01$ & $.84 \pm .01$ & $.84 \pm .01$\\ 
classic & $.94 \pm .01$ & $.94 \pm .01$ & $.94 \pm .01$ & $.94 \pm .01$ & $.94 \pm .00$ & $.94 \pm .00$ & $.94 \pm .00$ & $.94 \pm .00$ & $.94 \pm .00$ & $.94 \pm .00$ & $.94 \pm .00$\\ 
k1b & $.94 \pm .02$ & $.94 \pm .01$ & $.94 \pm .01$ & $.94 \pm .01$ & $.95 \pm .01$ & $.95 \pm .01$ & $.95 \pm .01$ & $.95 \pm .01$ & $.94 \pm .01$ & $.94 \pm .01$ & $.94 \pm .01$\\ 
hitech & $.61 \pm .04$ & $.61 \pm .03$ & $.61 \pm .03$ & $.61 \pm .03$ & $.60 \pm .02$ & $.60 \pm .02$ & $.60 \pm .02$ & $.60 \pm .02$ & $.60 \pm .02$ & $.60 \pm .02$ & $.60 \pm .02$\\ 
reviews & $.92 \pm .01$ & $.91 \pm .01$ & $.91 \pm .01$ & $.91 \pm .01$ & $.91 \pm .01$ & $.91 \pm .01$ & $.91 \pm .01$ & $.91 \pm .01$ & $.91 \pm .01$ & $.91 \pm .01$ & $.91 \pm .01$\\ 
sports & $.95 \pm .01$ & $.95 \pm .01$ & $.95 \pm .01$ & $.95 \pm .01$ & $.95 \pm .01$ & $.95 \pm .01$ & $.95 \pm .01$ & $.95 \pm .01$ & $.95 \pm .01$ & $.95 \pm .01$ & $.95 \pm .01$\\ 
la1 & $.82 \pm .03$ & $.82 \pm .03$ & $.82 \pm .02$ & $.82 \pm .02$ & $.82 \pm .02$ & $.82 \pm .02$ & $.82 \pm .02$ & $.82 \pm .02$ & $.82 \pm .01$ & $.82 \pm .01$ & $.82 \pm .01$\\ 
la12 & $.85 \pm .02$ & $.84 \pm .01$ & $.84 \pm .01$ & $.84 \pm .01$ & $.84 \pm .00$ & $.84 \pm .01$ & $.84 \pm .00$ & $.84 \pm .01$ & $.84 \pm .01$ & $.84 \pm .01$ & $.84 \pm .00$\\ 
la2 & $.83 \pm .02$ & $.84 \pm .01$ & $.84 \pm .01$ & $.84 \pm .01$ & $.84 \pm .01$ & $.84 \pm .01$ & $.84 \pm .01$ & $.84 \pm .01$ & $.84 \pm .01$ & $.84 \pm .01$ & $.84 \pm .01$\\ 
tr11 & $.72 \pm .07$ & $.72 \pm .08$ & $.71 \pm .08$ & $.70 \pm .07$ & $.69 \pm .07$ & $.69 \pm .06$ & $.69 \pm .06$ & $.69 \pm .06$ & $.69 \pm .06$ & $.69 \pm .06$ & $.69 \pm .06$\\ 
tr23 & $.73 \pm .08$ & $.71 \pm .08$ & $.69 \pm .08$ & $.69 \pm .07$ & $.69 \pm .07$ & $.68 \pm .07$ & $.68 \pm .07$ & $.68 \pm .07$ & $.68 \pm .07$ & $.68 \pm .07$ & $.68 \pm .07$\\ 
tr41 & $.90 \pm .04$ & $.90 \pm .03$ & $.90 \pm .03$ & $.90 \pm .03$ & $.90 \pm .02$ & $.90 \pm .02$ & $.90 \pm .02$ & $.90 \pm .02$ & $.90 \pm .02$ & $.90 \pm .02$ & $.90 \pm .02$\\ 
tr45 & $.80 \pm .04$ & $.81 \pm .04$ & $.82 \pm .04$ & $.82 \pm .04$ & $.82 \pm .04$ & $.82 \pm .03$ & $.82 \pm .04$ & $.82 \pm .03$ & $.82 \pm .03$ & $.82 \pm .04$ & $.82 \pm .04$
\\ \hline \end{tabular}}
\vspace{12pt} \caption{Results as AC for all the
datasets. Each column indicates the results obtained including the corresponding
fold in the test set.}\label{tab:streamingAcc}
\end{table}%

The results of this evaluation are presented in Table \ref{tab:streamingMI} and
\ref{tab:streamingAcc} as NMI and AC, respectively. From the tables we can see
that the performances of the system are stable over time. In fact, we can see
that in $9$ datasets over $13$, the different among the results as NMI with the
entire dataset ($12$ folds) and those with $2$ folds is $~2\%$. The results as
AC are even better. In fact, with the entire dataset the performances are stable
and in two cases higher (\emph{la2} and \emph{tr45}). The latter behavior can be
explained considering the fact that the algorithm exploit contextual information
and in many cases having more information to use leads to better solutions. We
can see that just in one case we have a drop of $5\%$ in performances, comparing
the results in fold $2$ with those in fold $12$. The most negative results have
been achieved on small datasets, this because in these cases the clusters are
small and unbalanced. In particular dealing with clusters of very different
sizes makes the $k$-nn algorithm, used to sparsify the graph, not useful. In
fact, the resulting structure allow the elements of small clusters to have
connections with elements belonging to other clusters. In these cases the
dynamics of our system converge to solutions in which small clusters are
absorbed by bigger ones. This because the elements belonging to small clusters
are likely to receive influence from the elements belonging to large clusters if
$k$ is larger than the cardinality of the small clusters. This phenomenon can be 
seen in Fig. \ref{fig:streaming}, where we compare the clustering results of our 
method against the ground truth, on \emph{k1b}. We can see that the orange 
cluster disappears in fold $2$ and that this error is propagated on the other 
folds. The other clusters are partitioned correctly.

If we compare the results in this Section with the results proposed in Section
\ref{sec:compsota} we can see that with this approach we can have a bust in
performances. In fact, in all datasets, except one (tr11) the results are higher
both in terms of NMI and AC. We can see that using just few labeled points
allows our approach to substantially improve its performances. Furthermore we
see that these performance are stable over time and that the standard deviation
is very low in all experiments, $\leq 0.11$ for NMI and $\leq 0.8$ for AC.

\subsubsection{Comparison with k-nn}%
We conducted the same experiment described in previous Section to compare the
performances of our method with the k-nearest neighbor (k-NN) algorithm. We used
k-NN to classify iteratively the folds of each dataset treating the data in same
way of previous experiments and setting $k=1$. Experimentally we noticed that
this value achieve the best performances. Higher values have very low NMI,
leading to situations in which small clusters are merged in bigger ones.

\begin{table}
  \resizebox{\textwidth}{!}{%
\begin{tabular}{| l | c | c | c | c | c | c | c | c | c | c | c |} \hline
Data & 2 & 3 & 4 & 5 & 6 & 7 & 8 & 9 & 10 & 11 & 12  \\ \hline
ng3sim & $.25 \pm .03$ & $.31 \pm .03$ & $.36 \pm .03$ & $.40 \pm .02$ & $.43 \pm .02$ & $.46 \pm .01$ & $.48 \pm .01$ & $.49 \pm .01$ & $.51 \pm .01$ & $.52 \pm .01$ & $.53 \pm .01$\\ 
classic & $.31 \pm .02$ & $.39 \pm .02$ & $.44 \pm .02$ & $.49 \pm .02$ & $.52 \pm .01$ & $.55 \pm .01$ & $.58 \pm .01$ & $.60 \pm .01$ & $.62 \pm .01$ & $.63 \pm .01$ & $.64 \pm .01$\\ 
k1b & $.32 \pm .04$ & $.38 \pm .03$ & $.44 \pm .02$ & $.49 \pm .02$ & $.53 \pm .02$ & $.57 \pm .02$ & $.60 \pm .01$ & $.62 \pm .01$ & $.64 \pm .01$ & $.66 \pm .01$ & $.67 \pm .01$\\ 
hitech & $.17 \pm .03$ & $.18 \pm .02$ & $.20 \pm .02$ & $.21 \pm .02$ & $.23 \pm .02$ & $.24 \pm .01$ & $.26 \pm .01$ & $.27 \pm .01$ & $.28 \pm .01$ & $.29 \pm .01$ & $.29 \pm .01$\\ 
reviews & $.35 \pm .03$ & $.41 \pm .03$ & $.46 \pm .02$ & $.50 \pm .02$ & $.53 \pm .02$ & $.55 \pm .01$ & $.57 \pm .01$ & $.59 \pm .01$ & $.60 \pm .01$ & $.61 \pm .01$ & $.62 \pm .01$\\ 
sports & $.48 \pm .02$ & $.56 \pm .02$ & $.62 \pm .01$ & $.66 \pm .01$ & $.69 \pm .01$ & $.71 \pm .01$ & $.73 \pm .01$ & $.75 \pm .01$ & $.76 \pm .01$ & $.77 \pm .00$ & $.78 \pm .00$\\ 
la1 & $.31 \pm .02$ & $.35 \pm .02$ & $.39 \pm .02$ & $.42 \pm .02$ & $.44 \pm .02$ & $.46 \pm .02$ & $.48 \pm .01$ & $.50 \pm .01$ & $.51 \pm .01$ & $.52 \pm .01$ & $.53 \pm .01$\\ 
la12 & $.32 \pm .02$ & $.37 \pm .02$ & $.41 \pm .01$ & $.45 \pm .01$ & $.48 \pm .01$ & $.50 \pm .01$ & $.52 \pm .01$ & $.53 \pm .01$ & $.55 \pm .01$ & $.56 \pm .01$ & $.57 \pm .01$\\ 
la2 & $.33 \pm .03$ & $.37 \pm .03$ & $.41 \pm .02$ & $.44 \pm .02$ & $.47 \pm .01$ & $.49 \pm .01$ & $.51 \pm .01$ & $.53 \pm .01$ & $.54 \pm .01$ & $.55 \pm .01$ & $.56 \pm .01$\\ 
tr11 & $.36 \pm .07$ & $.38 \pm .04$ & $.40 \pm .04$ & $.43 \pm .04$ & $.45 \pm .03$ & $.47 \pm .03$ & $.49 \pm .03$ & $.50 \pm .02$ & $.52 \pm .02$ & $.53 \pm .02$ & $.54 \pm .02$\\ 
tr23 & $.34 \pm .12$ & $.35 \pm .09$ & $.39 \pm .06$ & $.40 \pm .06$ & $.41 \pm .07$ & $.44 \pm .06$ & $.46 \pm .06$ & $.47 \pm .05$ & $.49 \pm .04$ & $.50 \pm .04$ & $.52 \pm .04$\\ 
tr41 & $.41 \pm .05$ & $.47 \pm .04$ & $.51 \pm .03$ & $.55 \pm .03$ & $.59 \pm .02$ & $.61 \pm .02$ & $.63 \pm .02$ & $.65 \pm .02$ & $.67 \pm .02$ & $.68 \pm .02$ & $.70 \pm .01$\\ 
tr45 & $.46 \pm .05$ & $.48 \pm .05$ & $.52 \pm .04$ & $.55 \pm .03$ & $.57 \pm .02$ & $.60 \pm .02$ & $.62 \pm .02$ & $.63 \pm .02$ & $.64 \pm .02$ & $.65 \pm .01$ & $.66 \pm .01$ 
\\ \hline
\end{tabular}}\vspace{12pt}
\caption{Results as NMI for all the datasets using k-NN.
Each column indicates the results obtained including the corresponding fold in the test set.}
\label{tab:streamingMIknn}
\end{table}

\begin{table}
  \resizebox{\textwidth}{!}{%
\begin{tabular}{| l | c | c | c | c | c | c | c | c | c | c | c | c |} \hline
Data & 2 & 3 & 4 & 5 & 6 & 7 & 8 & 9 & 10 & 11 & 12  \\ \hline
ng3sim & $.60 \pm .02$ & $.67 \pm .02$ & $.72 \pm .01$ & $.76 \pm .01$ & $.78 \pm .01$ & $.80 \pm .01$ & $.81 \pm .01$ & $.82 \pm .01$ & $.83 \pm .01$ & $.84 \pm .01$ & $.84 \pm .00$\\ 
classic & $.59 \pm .02$ & $.68 \pm .01$ & $.73 \pm .01$ & $.77 \pm .01$ & $.80 \pm .01$ & $.82 \pm .01$ & $.84 \pm .01$ & $.85 \pm .00$ & $.86 \pm .00$ & $.87 \pm .00$ & $.87 \pm .00$\\ 
k1b & $.53 \pm .04$ & $.62 \pm .02$ & $.69 \pm .02$ & $.74 \pm .01$ & $.78 \pm .01$ & $.81 \pm .01$ & $.83 \pm .01$ & $.84 \pm .01$ & $.86 \pm .01$ & $.87 \pm .01$ & $.88 \pm .01$\\ 
hitech & $.40 \pm .03$ & $.44 \pm .02$ & $.48 \pm .02$ & $.51 \pm .02$ & $.53 \pm .01$ & $.55 \pm .01$ & $.57 \pm .01$ & $.58 \pm .01$ & $.59 \pm .01$ & $.60 \pm .01$ & $.61 \pm .01$\\ 
reviews & $.58 \pm .03$ & $.66 \pm .02$ & $.72 \pm .01$ & $.76 \pm .01$ & $.78 \pm .01$ & $.81 \pm .01$ & $.82 \pm .01$ & $.83 \pm .00$ & $.84 \pm .00$ & $.85 \pm .00$ & $.86 \pm .00$\\ 
sports & $.72 \pm .01$ & $.79 \pm .01$ & $.83 \pm .01$ & $.86 \pm .01$ & $.88 \pm .00$ & $.89 \pm .00$ & $.90 \pm .00$ & $.91 \pm .00$ & $.92 \pm .00$ & $.92 \pm .00$ & $.93 \pm .00$\\ 
la1 & $.46 \pm .02$ & $.55 \pm .01$ & $.61 \pm .01$ & $.66 \pm .01$ & $.69 \pm .01$ & $.71 \pm .01$ & $.73 \pm .01$ & $.74 \pm .01$ & $.76 \pm .01$ & $.77 \pm .01$ & $.78 \pm .01$\\ 
la12 & $.49 \pm .01$ & $.58 \pm .01$ & $.64 \pm .01$ & $.68 \pm .01$ & $.72 \pm .01$ & $.74 \pm .01$ & $.76 \pm .01$ & $.77 \pm .01$ & $.78 \pm .01$ & $.79 \pm .00$ & $.80 \pm .00$\\ 
la2 & $.49 \pm .03$ & $.58 \pm .02$ & $.64 \pm .02$ & $.68 \pm .01$ & $.71 \pm .01$ & $.73 \pm .01$ & $.75 \pm .01$ & $.76 \pm .01$ & $.78 \pm .01$ & $.78 \pm .00$ & $.79 \pm .00$\\ 
tr11 & $.42 \pm .05$ & $.43 \pm .04$ & $.46 \pm .04$ & $.50 \pm .04$ & $.55 \pm .03$ & $.58 \pm .03$ & $.61 \pm .02$ & $.63 \pm .02$ & $.66 \pm .02$ & $.67 \pm .02$ & $.69 \pm .02$\\ 
tr23 & $.49 \pm .07$ & $.49 \pm .05$ & $.54 \pm .04$ & $.59 \pm .04$ & $.63 \pm .04$ & $.66 \pm .04$ & $.69 \pm .04$ & $.71 \pm .03$ & $.73 \pm .03$ & $.75 \pm .03$ & $.76 \pm .03$\\ 
tr41 & $.45 \pm .05$ & $.50 \pm .03$ & $.55 \pm .02$ & $.62 \pm .02$ & $.67 \pm .02$ & $.71 \pm .01$ & $.73 \pm .01$ & $.76 \pm .01$ & $.78 \pm .01$ & $.80 \pm .01$ & $.81 \pm .01$\\ 
tr45 & $.50 \pm .04$ & $.56 \pm .04$ & $.62 \pm .03$ & $.67 \pm .02$ & $.71 \pm .02$ & $.74 \pm .01$ & $.76 \pm .01$ & $.78 \pm .01$ & $.79 \pm .01$ & $.80 \pm .01$ & $.81 \pm .01$
\\ \hline \end{tabular}}\vspace{12pt} \caption{Results as AC for all the
datasets using k-NN. Each column indicates the results obtained including the corresponding
fold in the test set.}\label{tab:streamingAccknn}
\end{table}%

The results of this evaluation are shown in Table \ref{tab:streamingMIknn} and
\ref{tab:streamingAccknn} as NMI and AC, respectivelly. From these tables we can
see that the performances of k-NN are not stable and tend to increase at each
step. We can notice that the results in fold $2$ in many cases are doubled in 
fold $12$, this behaviour demonstrate that this algorithm requires many data to 
achieve good classification performances. To the contrary with our approach it 
is possible to obtain stable performances in each fold.

The performances of k-NN are very low compared with our approaches. In
particular, we can see that it does not perform well in the first seven folds.
This can be explained considering that it classify new instances taking into
account only local information (the information on the class membership of its
nearest neighbour), without considering any other source of information and
without imposing any coherence constraint using contextual information.

Form Table \ref{tab:streamingMIknn} and \ref{tab:streamingAccknn} we can see
that the results of k-NN in fold $12$ (entire dataset) are almost always lower
that those obtained with our method, both in terms of NMI and AC. In fact, just
in two cases k-NN obtain equal and higher results, in \emph{tr11} and
\emph{tr23} if we consider the NMI. If we consider the results as AC we can see
that in two datasets k-NN has the same performances of our method (\emph{NG17-19}
and \emph{tr11}) and that it has higher performances on \emph{hitech} ($+1\%$).

\section{Conclusions}%
With this work we explored new methods for document clustering based on game
theory and consistent labeling principles. We have conducted an extensive series
of experiments to test the approach on different scenarios. We have also
evaluated the system with different implementations and compared the results
with state-of-the-art algorithms. 

Our method can be considered as a continuation of graph based approaches but it
combines together the partition of the graph and the propagation of the
information across the network. With this method we used the structural
information about the graph and then we employed evolutionary dynamics to find
the best labeling of the data points. The application of a game theoretic
framework is able to exploit relational and contextual information and
guarantees that the final labeling of the data is consistent.

The system has demonstrated to perform well compared with state-of-the-art
system and to be extremely flexible. In fact, it has been tested with different
features, with and without the information about the number of clusters to
extract and on static and dynamic context. Furthermore, it is not difficult to
implement new graph similarity measure and new dynamics to improve its
performances or to adapt to new contexts.

The experiments without the use of $K$, where the algorithm collects together
highly similar points and then merges the resulting groups, demonstrated how it
is able to extract clusters of any size without the definition of any centroid.
The experiments on streaming data demonstrated that our approach can be used to
cluster data dynamically. In fact, the performances of the system does not
change much when the test set is enriched with new instances to cluster. This is
an appealing feature, since it makes the framework flexible and not
computationally expensive. On this scenario it was demonstrated that the use of
contextual information helps the clustering task. In fact, using the k-NN
algorithm on streaming data produces lower and not stable results.

As future work we are planning to apply this framework to other kind of data and
also to use it in the context of \emph{big data}, where, in many cases, it is
necessary to deal with datasets that do not fit in memory and have to be divided
in different parts in order to be clustered or classified.

\section*{Acknowledgments} This work was partly supported by the Samsung Global
Research Outreach Program.

\bibliographystyle{splncs03}
\bibliography{main}

\end{document}